\documentclass{article}

\usepackage{arxiv}

\usepackage{amsmath}
\usepackage[T1]{fontenc} 
\usepackage[utf8]{inputenc} 
\usepackage{natbib} 
\usepackage{nicefrac}
\usepackage{booktabs} 
\usepackage{microtype} 

\usepackage{url} 
\usepackage{doi} 
\usepackage{xcolor}
\usepackage{lipsum}
\usepackage{listings}
\usepackage{tabularx} 
\usepackage{graphicx}
\usepackage{orcidlink}
\usepackage{nccmath} 
\usepackage{amsfonts} 
\usepackage{algorithm}
\usepackage{algpseudocode}

\title{Next Generation Loss Function for Image Classification}

\newif\ifuniqueAffiliation
\uniqueAffiliationtrue

\ifuniqueAffiliation 
\author{Shakhnaz~Akhmedova\,\orcidlink{0000-0003-2927-1974}\\
Centre for Artificial Intelligence in\\
Public Health Research\\
Robert Koch Institute\\
Berlin, Germany\\
\texttt{AkhmedovaS@rki.de} \\
\And
Nils K\"orber\\
Centre for Artificial Intelligence in\\ 
Public Health Research\\
Robert Koch Institute\\
Berlin, Germany\\
\texttt{KoerberN@rki.de}\\
}

\begin{document}
\maketitle

\begin{abstract}
Neural networks are trained by minimizing a loss function that defines the discrepancy between the predicted model output and the target value. The selection of the loss function is crucial to achieve task-specific behaviour and highly influences the capability of the model. A variety of loss functions have been proposed for a wide range of tasks affecting training and model performance. For classification tasks, the cross entropy is the de-facto standard and usually the first choice. Here, we try to experimentally challenge the well-known loss functions, including cross entropy (CE) loss, by utilizing the genetic programming (GP) approach, a population-based evolutionary algorithm. GP constructs loss functions from a set of operators and leaf nodes and these functions are repeatedly recombined and mutated to find an optimal structure. Experiments were carried out on different small-sized datasets CIFAR-10, CIFAR-100 and Fashion-MNIST using an Inception model. The 5 best functions found were evaluated for different model architectures on a set of standard datasets ranging from 2 to 102 classes and very different sizes. One function, denoted as Next Generation Loss (NGL), clearly stood out showing same or better performance for all tested datasets compared to CE. To evaluate the NGL function on a large-scale dataset, we tested its performance on the Imagenet-1k dataset where it showed improved top-1 accuracy compared to models trained with identical settings and other losses. Finally, the NGL was trained on a segmentation downstream task for Pascal VOC 2012 and COCO-Stuff164k datasets improving the underlying model performance.

\end{abstract}

\keywords{Loss function \and Genetic Programming \and Deep Learning \and Classification \and Segmentation.}

\section{Introduction}
\begin{figure}
\centering
\includegraphics[width=\textwidth]{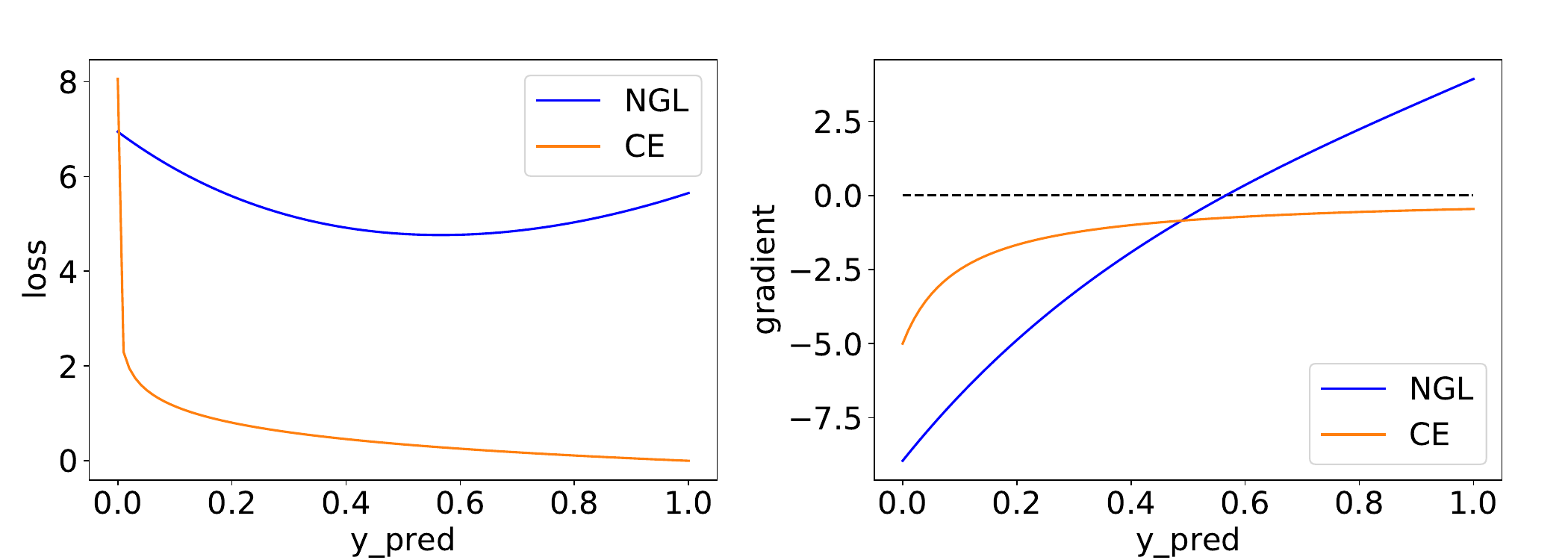}
\caption{NGL and CE functions (left) and their gradients (right) for $y_{real}=1$.} \label{fig_loss_graphs}
\end{figure}

Deep learning (DL) has received extensive research interest in developing new image processing algorithms, and resulting models have been remarkably successful in a variety of image analysis tasks such as classification~\cite{krizhevsky2012imagenet}, segmentation~\cite{Ronneberger2015UNetCN}, object detection~\cite{girshick2015fast} and more. Deep learning methods represented by (or partially based on) convolutional neural networks have become a well-known and constantly developing research topic in the field of image analysis. Various deep learning models, for example ResNet~\cite{He2015DeepRL}, Inception~\cite{Szegedy2015RethinkingTI}, U-Net~\cite{Ronneberger2015UNetCN}, DeepLab~\cite{Chen2016DeepLabSI}, have been proposed over the years for different image processing tasks and domains. Vision transformers~\cite{Dosovitskiy2020AnII} constantly gain popularity among researchers due to their high training efficiency and scalability and are already outperforming convolutional neural networks on multiple benchmarks for classification, segmentation and image generation problems.

The loss function plays a crucial role in deep learning in general, as it shows the direction for model training and affects the performance of the model and training speed. There are well-known losses used for most image processing tasks, such as cross entropy (CE)~\cite{Cover2006ElementsOI}, Focal Loss~\cite{lin2017focal}, or mean square error~\cite{Willmott2005AdvantagesOT}. Recently, researchers started proposing loss functions developed for specific applications or tasks. Usually, that is done by hand and requires a lot of expert knowledge, time, parameter adjustments, and thus computational resources, due to the complexity of modern models.

The loss function design can be represented as a symbolic regression problem that can be solved by using Genetic Programming (GP)~\cite{Augusto00}. The GP approach is frequently used to design previously unknown heuristics in order to solve hard computational search problems~\cite{Burke09}. Genetic programming is a domain-independent method, which was used in this study to genetically breed a population of loss functions for deep learning models to solve image classification problems. GP was applied to find a generally applicable loss function for DL classifiers independent of the dataset and model architecture. Several experiments with different objective functions using InceptionV3~\cite{Szegedy2015RethinkingTI} trained on the CIFAR-10~\cite{krizhevsky2009learning}, Fashion-MNIST~\cite{xiao2017/online} and CIFAR-100~\cite{krizhevsky2009learning} datasets were conducted to find an optimized loss function. 

The loss functions found were evaluated using ResNet50 and InceptionV3 models trained on CIFAR-10~\cite{krizhevsky2009learning}, CIFAR-100~\cite{krizhevsky2009learning}, Fashion-MNIST~\cite{xiao2017/online}, Malaria~\cite{rajaraman2018pre}, PCam~\cite{Veeling2018-qh}, Colorectal Histology~\cite{kather2016multi} and Caltech 101~\cite{FeiFei2004LearningGV} datasets. In the next step, the best function, denoted as NGL, was used to train ResNet~\cite{He2015DeepRL} models and Swin transformers~\cite{liu2021Swin} to classify the ImageNet-1k dataset. In the end, it was applied to train U-Net and DeepLab models on Pascal VOC 2012~\cite{pascal-voc-2012} and COCO-Stuff164k~\cite{caesar2018cvpr} datasets to show the applicability for segmentation.

Among the evaluated functions, NGL demonstrated exceptionally good results outperforming baseline losses, such as cross entropy loss, focal loss, symmetric cross entropy and dice loss. Moreover, further investigation has showed that the reason for these good results may be self-regulation, which is inherently present in the NGL due to its mathematical definition.

Thus, the main contributions in this paper can be summarized as follows:
\begin{itemize}
\item Experimental results show that GP is capable to design a good loss function for deep neural networks without any prior knowledge neither about the model nor about the problem at hand;
\item NGL, a new highly efficient loss function, applicable for any DL model to solve classification and segmentation problems, was found;
\item Self-regularization naturally implemented in the NGL indicates a direction for loss function design in general;
\item Models trained by the NGL achieved same or better performance on a variety of datasets including ImageNet-1k and COCO-Stuff164k datasets.
\end{itemize}

The rest of this paper is organized as follows. Section 2 briefly reviews related work of deep neural networks used for image analysis and loss function design. The Genetic Programming algorithm and proposed search procedure are described in Section 3, while the experimental settings and results are presented in Section 4. Obtained results are discussed in Section 5. Finally, the paper is concluded in Section 6.

\section{Related Work}
The progress in solving image processing and visual recognition tasks is related to the rapid development of convolutional neural networks (CNNs) and subsequently vision transformers, which have been already used for real-world applications in different scientific fields. A plethora of convolutional neural network architectures have been proposed, including the widely used Inception network~\cite{Szegedy2014GoingDW} that explores the problem of multi-scale fusion in convolution calculations to better characterize image information, and ResNet~\cite{He2015DeepRL} with residual blocks to solve the vanishing gradients problem. 

Dosovitskiy et al. ~\cite{Dosovitskiy2020AnII} proposed the Vision Transformer (ViT) by formulating image classification as a sequence prediction task of the image patch sequence similarly to ~\cite{Vaswani2017AttentionIA}, thereby capturing long-term dependencies within the input image. Transformer-based models for image processing have been rapidly developed, and the following models are considered as some of the most representative: Vision Transformers (ViT)~\cite{Dosovitskiy2020AnII}, Data-efficient image Transformers (DeiT)~\cite{Touvron2020TrainingDI} and Swin-Transformer~\cite{liu2021Swin}.

Most deep learning models initially designed to solve classification problems can be applied also for segmentation tasks, as segmentation can be formulated as a classification problem of pixels with semantic labels (semantic segmentation) or partitioning of individual objects (instance segmentation). In the recent decade there were a lot of works presenting new and more powerful segmentation methods, U-Net~\cite{Ronneberger2015UNetCN}, which was initially developed for biomedical image segmentation, is one of the most widely used architectures not only in the life sciences. Additionally, DeepLabv1~\cite{Chen2016DeepLabSI}, DeepLabv2~\cite{Chen2016DeepLabSI} and DeepLabv3+~\cite{deeplabv3plus2018} are among some of the most popular image segmentation approaches, which use dilated convolution to address the decreasing resolution in the network (caused by max-pooling and striding) and Atrous Spatial Pyramid Pooling (ASPP), that probes an incoming convolutional feature layer with filters at multiple sampling rates.

Aside from developing deeper networks with more complex structures and features to get better performance, better loss functions have also been proven to be effective on improving the model performance in most recent works~\cite{Tian2022RecentAO},~\cite{Jadon2020ASO}. Many loss functions have been utilized for neural networks based on softmax activation function, such as Mean Square Error (MSE) loss function~\cite{Willmott2005AdvantagesOT}, Cross Entropy (CE) loss function~\cite{Cover2006ElementsOI} (Binary Cross Entropy loss function when images are divided into two classes, Categorical Cross Entropy for multiple classes), etc. In most state-of-the-art supervised learning problems, practitioners typically use large capacity deep neural networks together with cross-entropy loss. The latter can be explained by the fact that the traditional CE loss is supported by clear theory, easy training and good performance. Nowadays, there are many variants of CE that have been proposed in the past few years, e.g. symmetric CE (SCE)~\cite{Han2017ANA}, or Focal Loss~\cite{lin2017focal}.

Loss functions used for segmentation can be divided into four groups described in~\cite{Jadon2020ASO}: distribution-based, region-based, boundary-based, and compounded losses. The most well-known and commonly used distribution-based loss functions used to train segmentation models are binary cross entropy~\cite{Yide2004AutomatedIS} and focal loss~\cite{Lin2017FocalLF}. Region-based loss functions such as dice loss~\cite{Sudre2017GeneralisedDO}, and Tversky loss~\cite{Salehi2017TverskyLF} calculate the similarity between images. Hausdorff distance loss~\cite{Ribera2018WeightedHD} is defined as boundary-based loss functions in~\cite{Jadon2020ASO}, while the exponential logarithmic loss~\cite{Wong20183DSW} is considered as the compounded loss.

Most of the mentioned loss functions can be considered as heuristic methods designed by experts by using theory and the domain knowledge for specific tasks and/or models. Moreover, loss function design requires great effort from experts to explore the large search space, which is usually sub-optimal in practice. Nevertheless, only several commonly used losses are usually implemented for model training, for example, categorical cross entropy for classification or dice loss for segmentation. Recently, the automated search of suitable loss functions without domain knowledge has received much attention from the computer vision community. Reinforcement learning algorithms were used to learn better loss functions with good generalization ability on different image analysis tasks~\cite{Xu2018MetaGradientRL},~\cite{Oh2020DiscoveringRL}. However, loss functions found by using these approaches either have task-specific requirements, such as environment interaction in reinforcement learning, or remain fixed after training. 

A lot of works on automatic loss search follow the search algorithms used in AutoML~\cite{He2019AutoMLAS}. In~\cite{Li2021AutoLossZeroSL} the authors explore the possibility of searching loss functions automatically from scratch for generic tasks, e.g., semantic and instance segmentation, object detection, and pose estimation. In studies such as~\cite{Wang2020LossFS} and~\cite{Li2019AMLFSAF} the main focus is on the search for particular hyper-parameters within the fixed loss formula. AutoML-Zero~\cite{Real2020AutoMLZeroEM} proposes a framework to construct machine learning algorithms from simple mathematical operations, but the search space and search strategies are specialized, which limits its potential application. The authors of the study presented in~\cite{Gonzalez2019ImprovedTS} propose a framework for searching classification loss functions by using evolutionary algorithms but the searched loss poorly generalizes to large-scale datasets. These works motivated us to design loss functions by using the Genetic Programming algorithm, which uses a large set of primitive mathematical operations and terminal values, capable of training different models regardless of their structure or the task at hand.

\section{Method}
Genetic Programming (GP) is a population-based evolutionary optimization algorithm, which can be applied to design different heuristics, depending on the problem at hand, and, additionally, it is considered as a machine learning tool, as it can be used to discover a functional relationship between features in data. GP was initially inspired by the biological evolution, including natural processes such as mutation and selection. In GP, each solution is represented by a tree structure, and each of these trees are evaluated recursively to produce the resulting multivariate expression. There are two types of nodes used for tree-based GP: the terminal node, also called leaf, which is randomly chosen from the set of variables, and the tree node, which can be chosen from a predefined set of operators. An example of the generated trees is given in Appendix A (Figure~\ref{fig_GP}).
The GP search process starts with random initialization of a set of potential solutions, which is also called a population of individuals, in the functional search space. The number of individuals is the predefined parameter $n$, which does not change through the whole search process. The overall GP search process is summarized in Algorithm~\ref{alg_1}. Thus, for this study a set of loss functions was randomly generated, and each loss function was represented as a tree, where the terminals were randomly chosen from the set $\{y_{pred}, y_{real}, \mathbb{R}\}$, while the operators were chosen from the set $\{+, -, \times, \div, \times(-1), \sqrt, \log, \exp, \sin, \cos\}$. To utilize these functions and terminals for loss function generation, several modifications were applied (mentioned modifications are described in Appendix A).

\begin{algorithm}
\caption{An overview of the implemented GP approach}\label{alg_1}
\label{alg:alg_1}
\begin{algorithmic}
\State Randomly initialize $n$ trees (generation $0$), each representing one formula, e.g., loss function
\State Evaluate GP fitness function $F$ for each individual in the population
\State Determine the best individual
\State Create an empty external archive $A$ 
\While{generation number is less than $T$}
\For{each individual from population}
\State Generate a child individual by applying the crossover operator
\If{$rand_1<M_{ST}$}
\State Apply subtree mutation to the generated child
\EndIf
\If{$rand_2<M_N$}
\State Apply one-point mutation to the generated child
\EndIf
\EndFor
\State Evaluate GP fitness function $F$ for each generated child individual
\State Update the best individual
\State Create new population of size $2 \times n$ by concatenating children and parent individuals
\State Select $n$ best individuals from the new population: create new generation
\State Update the external archive $A$
\EndWhile
\end{algorithmic}
\end{algorithm}

After initialization, the main search loop starts, which consists of the iteratively repeated steps crossover, mutation, fitness function evaluation and selection (the number of GP steps or generations is denoted as $T$). The fitness function $F$ determines the quality of the individual and is the crucial part of the optimization process, as our goal was to experimentally find a robust loss function that can be generalized to a variety of datasets. 

The crossover operator is used to exchange the subtrees between two individuals. An example of the implemented crossover operator is demonstrated in Appendix A (Figure~\ref{fig_GP}). Mutation can be applied in various ways, but in this study two variants were used: 
\begin{itemize}
\item a random subtree in the tree is chosen and replaced with another randomly generated subtree (subtee mutation);
\item a random node in the tree is chosen and replaced with another randomly generated node (one-point mutation).
\end{itemize}

Both mutation operators as well as crossover have their own parameters, such as subtree and node mutation $M_{ST}$, $M_N$ and crossover $Cr$ rates, which determine how often individuals will be changed during the main search loop. Finally, all individuals and the ones generated after crossover and mutation steps are combined into one population, and then the selection operator is applied. In this study only $n$ most fit individuals from the combined population are chosen for the next generation on each GP step. 

Moreover, the success-history-based adaptation strategy was used to improve the efficiency of the GP approach. To be more specific, at the beginning of the main search loop an empty external archive $A$ was created. The maximum size of this archive was set equal to the population size, $n_A = n$. All individuals not passing selection could have been saved in the archive with a given probability $p_A$. Individuals saved in the external set were used during crossover step with some probability $Cr_A$.

\section{Experiments}
\subsection{Loss function search}
As mentioned in the previous section, loss function search was performed by the GP algorithm and essentially can be described as an optimization process. A set of individuals, where each individual is a mathematical formula representing a loss function, is generated. These individuals change by crossover and mutation operators and then a new set of loss functions is selected. These actions are repeated a given number of times, called generations, and in the end the best individual or loss function is determined. In this implementation, the crucial part is selection, which in turn depends on how the fitness of the loss function is defined, since only the fittest individuals are transferred over to the next generation.

In this study, five experiments were conducted to search the loss function by GP, which only differed in how the fitness function $F$ was defined. Specifically, the following definitions of $F$ were used for each individual loss for each experiment:

\begin{itemize}
\item Train the InceptionV3 model from scratch one time on CIFAR-10 dataset, the validation error was used as the fitness value;
\item Train the InceptionV3 model from scratch 3 times on Fashion-MNIST dataset, the averaged validation error was used as the fitness value;
\item Fine-tune the pre-trained InceptionV3 model one time on CIFAR-10 dataset, the validation error was used as the fitness value;
\item Fine-tune the pre-trained InceptionV3 model 3 times on Fashion-MNIST dataset, the averaged validation error was used as the fitness value;
\item Train the InceptionV3 model from scratch on CIFAR-10, Fashion-MNIST and CIFAR-100 datasets (once each). The validation error for each dataset was compared to the respective validation error obtained by the same model trained using CE loss, and individual fitness was represented by the pair of numbers indicating the number of wins and the percentage of improvement compared to CE.
\end{itemize}
It should be noted that a loss function was discarded if its values were not in the range $[10^{-5}, 10^5]$ and a new one was generated instead of it. The fitness function evaluation procedure is described in details in Appendix B.1 (Algorithm~\ref{alg_2}).

The model and datasets were selected to limit the training time required per generation while maintaining a reasonable evaluation of the selected loss functions. Regardless of the experiment, a batch size of 128, data augmentation including horizontal flip, width and height shift ($0.1$ for both), zoom ($0.2$), Adam optimizer~\cite{Diederik15} with reduction of the learning rate on plateau, and 50 epochs for CIFAR-10 and CIFAR-100 and $35$ for Fashion-MNIST were used to train the network. The top layer of the InceptionV3 model was replaced by $9$ new layers, consisting of Flatten, BatchNormalization ($\times 3$)~\cite{Ioffe2015BatchNA}, Dense ($\times 3$), Dropout ($\times 2$)~\cite{Srivastava14}, and a final Softmax layer.

The following parameters were used for GP regardless of the experiment: $n=16$, $T=100$, $M_{ST}=0.3$, $M_N=0.1$, $Cr=0.7$, $p_A=Cr_A=0.5$, minimum tree height and maximum tree size were set to 2 and 100, respectively.

The best performing functions for each of the five GP experiments were evaluated for different models and multiple datasets, all functions are listed in Appendix B.1.

\subsection{Evaluation}
\subsubsection{Small datasets for classification}
The evaluation of the five found functions was performed by training ResNet50 and InceptionV3 on seven datasets, which differed by the number of images, classes, by the type of images (grayscale and RGB), and their sizes. The brief description of datasets used in this study is given in Table~\ref{tab_1} ($N$ is the number of classes). ResNet50 was added to the initial experiments, as all the found functions were evaluated on InceptionV3 model during the GP search to rule out model specific properties. The top layers of both InceptionV3 and ResNet50 networks were not included, in both cases instead $9$ new layers such as Flatten, BatchNormalization ($\times 3$), Dense ($\times 3$) and Dropout ($\times 2$) were added. 

\begin{table}
\caption{Brief description of the small datasets used for classification.}\label{tab_1}
\resizebox{\textwidth}{!}{\begin{tabular}{cccccc}
\toprule
Dataset & $N$ & Dataset size & Class size & Image size & Image type \\
\midrule
Malaria & $2$ & $27558$ & $13779$ & edge lengths of $40-400$ pixels & RGB \\
PCam & $2$ & $327680$ & $163840$ & $96\times96$ & RGB \\ 
Colorectal Histology & $8$ & $5\times10^3$ & $625$ & $150\times150$ & RGB \\
CIFAR-10 & $10$ & $60\times10^3$ & $6\times10^3$ & $32\times32$ & RGB \\ 
Fashion-MNIST & $10$ & $70\times10^3$ &$7\times10^3$ & $28\times28$ & Grayscale \\ 
CIFAR-100 & $100$ & $60\times10^3$ & $6\times10^2$ & $32\times32$ & RGB \\ 
Caltech 101 & $102$ & $9144$ & $40-800$ & edge lengths of $200-300$ pixels & RGB \\
\bottomrule
\end{tabular}}
\end{table}

Parameter settings and data preprocessing varied for all datasets but were identical for all tested loss functions, specifics are shown in Appendix B.2 (Table~\ref{tab_A1}). Classification accuracy on the test set after the last epoch was used as the evaluation metric. 

Both InceptionV3 and ResNet50 models were trained five times for each dataset. Obtained results were averaged for both models and compared to the averaged CE (with and without $L2$ regularization), focal and SCE losses for the same models. In the end, the percentage of accuracy improvement or decrease compared to CE was determined. A final value from the range $[-0.5, 0.5]$ was considered as same performing, a value $>+0.5 \%$ better performing and $<-0.5 \%$ performing worse compared to CE.

Results obtained for all evaluated loss functions used to train models from scratch ($f_1,...,f_5$) are given in Table~\ref{tab_3}, results obtained for pre-trained models are presented in Appendix B.2 (Table~\ref{tab_A2}). Loss function $f_3$ demonstrates the worst results compared to others. It showed particularly bad results on Malaria, Colorectal Histology, CIFAR-100, and Caltech 101 datasets, it should be noted that the last three mentioned datasets have less than $10^3$ examples of each present class. Interestingly, no function outperformed CE loss on Malaria, CIFAR-10, and Fashion-MNIST, while 6 out of 9 functions showed significantly better results on PCam dataset. 

\begin{table}
\caption{Average accuracy (\%) for ResNet50 and InceptionV3 each trained 5 times for each dataset for cross entropy (CE) and percentual change of the tested loss functions compared to CE, best result per dataset is shown in bold.}\label{tab_3}
\resizebox{\textwidth}{!}{\begin{tabular}{ccccccccc}
\toprule
Loss & Malaria & PCam & Colorectal Histology & CIFAR-10 & Fashion-MNIST & CIFAR-100 & Caltech 101 & Mean \\
\midrule
CE & 94.0 & 69.4 & 88.9 & 92.8 & 94.0 & 68.2 & 72.5 & $\pm 0$ \\
SCE & -0.06 & +0.62 & -0.45 & -3.66 & -2.42 & -0.80 & +2.71 & -0.58\\
Focal & \textbf{+0.34} & +1.03 & + 2.89 & -0.64 & -0.02 & -2.14 & -0.78 & +0.10\\
CE + $L2$ & +0.11 & -0.64 & +0.88 & \textbf{+0.32} & + 0.09 & +0.38 & +3.67 & +0.69\\
$f_1$ &-0.27 &+0.81 &\textbf{+3.27} &-0.34 &\textbf{+0.31} &-0.84 &-3.67 &-0.10\\
$f_2$ &-0.09 &+2.89 &-7.56 &-0.81 &+0.04 &+0.60 &-3.61 &-1.22\\
$f_3$ &-27.98 &+3.24 &-62.45 &-1.79 &-0.48 &-59.36 &-41.67 &-27.21\\
$f_4$ &-0.21 &-0.13 &-1.98 &-0.14 &0.00 &-0.32 &-6.49 &-1.32\\
$f_5 (NGL)$ &-0.27 &\textbf{+7.07} &+1.77 &+0.12 &+0.07 &\textbf{+1.01} &\textbf{+5.00} &\textbf{+2.11}\\
\bottomrule
\end{tabular}}
\end{table}

Overall, $f_5$ is the only function showing on average significantly better results than CE, outperforming it on four out of seven datasets with an improvement ranging from 1 to 7 $\%$, while there was no significant difference on the remaining three datasets. 

Given the overall good performance of the function found by an evolutionary method, we renamed function $f_5$ as Next Generation Loss (NGL) function, with the following formula:
\begin{equation}
f_{NGL}=\frac{1}{N}\sum_{i=1}^N{\left[e^{(\alpha-y^{(i)}_{pred}\cdot(1+y^{(i)}_{real}))}-\cos(\cos(\sin(y^{(i)}_{pred})))\right]},
\end{equation}
where $\alpha=2.4092$, $N$ is the number of classes.

To further test the generalizability of NGL, we tested its performance on ImageNet dataset and larger models.

\subsubsection{ImageNet-1k}
Experiments on ImageNet-1k dataset were performed using the ResNet~\cite{He2015DeepRL} convolutional neural networks and Swin~\cite{liu2021Swin} transformer models using NGL, CE, SCE and focal losses with the same settings.

For ResNet, we used the AdamW optimizer~\cite{loshchilov2019decoupled}, cosine annealing learning rate scheduler with a $30$-epoch linear warm-up, while the learning rate was equal to $0.003$. The number of epochs was set to $180$, a batch size of $256$, and a weight decay of $0.01$ were used. For Swin-Transformers, the AdamW optimizer was employed for $400$ epochs using a cosine decay learning rate scheduler and $60$ epochs of linear warm-up. A batch size of $1024$, an initial learning rate of $0.001$, and a weight decay of $0.05$ were used. All of the augmentation and regularization strategies were the same as in~\cite{liu2021Swin}. Results on ImageNet-1k are shown in Table~\ref{tab_4}. It should be noted, that for both ResNet and Swin models CE was applied with label smoothing and $L2$ regularization.

\begin{table}
\caption{Comparison of ImageNet results for models trained on CE, SCE NGL and focal losses. \dag denote results reported as 10-crop testing, all other results show single crop accuracy.}\label{tab_4}
\resizebox{\textwidth}{!}{\begin{tabular}{cccccc}
\toprule
Model & Original top1-acc & Retrained (CE) top1-acc & SCE top1-acc & Focal top1-acc & NGL top1-acc \\
\midrule
ResNet101 ~\cite{He2015DeepRL} & 78.25 \dag & 76.55 & 77.21 & 75.68 & 78.38\\
ResNet152 ~\cite{He2015DeepRL} & 78.57 \dag & 76.92 & 77.51 & 76.06 &78.99\\
Swin-T ~\cite{liu2021Swin} & 81.3 & 81.19 & 78.79 & 77.68 & 81.25\\
Swin-S ~\cite{liu2021Swin} & 83.0 & 83.1 & 80.68 & 79.16 & 83.0\\
\bottomrule
\end{tabular}}
\end{table}

For ResNet models NGL shows a clear improvement over other losses, increasing the top-1 accuracy by $1-3 \%$ . For the Swin architecture NGL significantly outperforms SCE and focal losses, while it shows similar performance as regularized CE. 

\begin{figure}
\centering
\includegraphics[width=\textwidth]{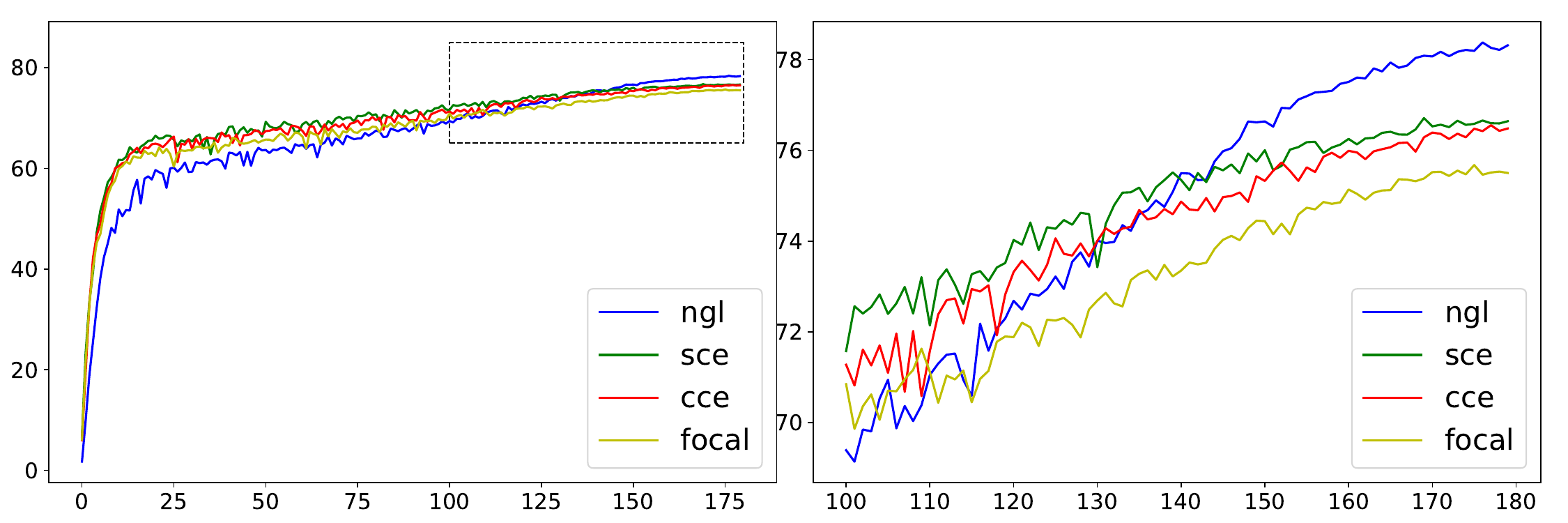}
\caption{Mean validation accuracy of ResNet101 model on each epoch during ImageNet-1k training.} \label{fig_acc_resnet101}
\end{figure}

The training process for ResNet101 model is demonstrated in Figure~\ref{fig_acc_resnet101}. It can be seen that NGL showed slower increase in accuracy compared to other losses, but converges later with a higher accuracy. Even though the NGL was found using only very small datasets, it shows astonishingly good performance on ImageNet-1k. For different architectures and model sizes NGL shows on average superior performance compared to mentioned loss functions.

\subsubsection{Segmentation}
Additionally, NGL was evaluated on segmentation problems. Specifically, it was used to train two segmentation models, DeepLabv2 with ResNet101 as backbone~\cite{nakashima2018deeplab} and U-Net with ResNet34 as backbone. The U-Net model was trained and evaluated only on Pascal VOC 2012~\cite{pascal-voc-2012} dataset, while the DeepLabv2 model was used for both Pascal VOC 2012 and COCO-Stuff164k~\cite{caesar2018cvpr} datasets. 

The first experiments were conducted utilizing the U-Net model. The same parameter setting was used for CE, focal, dice and NGL losses: images were resized to $224\times224$, Adam optimizer was applied, the learning rate was $0.0001$ and reduced on plateau with a factor of $0.2$, patience $5$ and $lr_{min}=10^{-6}$ was used to change it according to the IoU metric value on validation dataset. The model was trained for $100$ epochs with a batch size of $32$. For DeepLabv2, Adam optimizer was chosen, the learning rate $lr$ was set to $2.5\times10^{-5}$ and the polynomial learning rate scheduler with weight decay $5\times10^{-4}$ and power $0.9$ was applied to decrease $lr$ during $2\times10^4$ iterations. For both models the number of program runs was equal to $5$ and mean IoU value was used to evaluate the performance of the model.

Results for Pascal VOC are presented in Table~\ref{tab_5}. Using the U-Net shows approximately $3-7\%$ improvement in terms of mean IoU when NGL is used for training compared to other losses, while showing the same results as CE and ouperforming dice and focal losses using DeepLabv2. 

\begin{table}
\caption{Comparison of results (mIoU values) on Pascal VOC trained by CE, focal and dice and NGL loss functions.}\label{tab_5}
\centering
\begin{tabular}{ccccc}
\toprule
Model & CE & Focal & Dice & NGL \\
\midrule
U-Net & 50.1 & 49.1 & 46.1 & 52.8 \\
DeepLabv2 & 76.7 & 75.9 & 74.7 & 76.7\\
\bottomrule
\end{tabular}
\end{table}

Finally, the DeepLabv2 model was trained on COCO-Stuff164k using mentioned loss functions. The parameter setting for these experiments was the same as for Pascal VOC 2012 dataset with the maximum number of iterations set to $3\times10^5$. A comparison of results is shown in Table~\ref{tab_6}, showing slight improvement for NGL loss compared to CE, and significant improvement compared to focal and dice losses.

\begin{table}
\caption{Comparison of results achieved by DeepLabv2 model on COCO-Stuff164k dataset.}\label{tab_6}
\centering
\begin{tabular}{cccc}
\toprule
& Pixel Accuracy & Mean IoU & Mean Accuracy \\
\midrule
CE & $66.9$ & $39.8$ & $51.8$\\ 
Focal & $67.7$ & $39.5$ & $51.5$\\
Dice & $39.7$ & $17.3$ & $26.9$\\
NGL & $67.8$ & $39.9$ & $51.6$\\
\bottomrule
\end{tabular}
\end{table}

The obtained results demonstrate that NGL is suitable for downstream segmentation tasks and showing improved performance compared to other commonly used losses.

\section{Discussion}

For qualitative analysis, the NGL function can be simplified to a binary classification problem, similar as in ~\cite{Gonzalez2019ImprovedTS}. For a binary classification problem, NGL is as follows:
\begin{equation}
\begin{split}
f_{NGL} = \frac{1}{2}\left[e^{(\alpha-y^{(0)}_{pred} \cdot (1 + y_{real}))} + e^{(\alpha-(1-y^{(0)}_{pred}) \cdot (2 - y_{real}))}\right]\\
- \frac{1}{2}\left[\cos{(\cos{(\sin{(y^{(0)}_{pred})})})} + \cos{(\cos{(\sin{(1 - y^{(0)}_{pred})})})}\right] ,
\end{split}
\end{equation}
where $\alpha=2.4092$, $y_{pred}=(y^{(0)}_{pred}, y^{(1)}_{pred})$, $y^{(1)}_{pred}=1-y^{(0)}_{pred}$ and $y_{real}\in\{0,1\}$. 

Let us assume that the true labels $y_{real}$ are either $0$ or $1$. The case where $y_{real}=1$ is plotted in Figure~\ref{fig_loss_graphs} for the CE loss and NGL functions. The cross entropy shows a monotonic decrease with $y_{pred}$ converging to 1, the NGL shows a decrease in loss, resulting in a minimum at $0.57$ and slightly increases when $y_{pred}$ converges to 1. This increase in loss for $y_{pred}$ approaching the true value seems to be counter-intuitive, but may be a factor for the overall good performance of the loss function. The mentioned increase of the loss value may prevent the model from becoming too confident in its output predictions and may provide an important advantage, as it lowers the probability of overfitting. Thus, this could provide an implicit form of regularization, enabling better generalization. A more detailed explanation is given in Appendix C.

While some of the functions found during GP showed better performance on certain datasets, NGL was the only function that performed better than CE (with and without regularization), SCE, focal and dice losses on average and was therefore used for ImageNet-1k and COCO-Stuff164k training, where it was able to show its generalizability to larger datasets and models. The NGL is independent of additional parameters, it is differentiable and has an implicit regularization. For larger datasets it could be observed that it converges slower than other mentioned losses, but at a higher maximum accuracy. The NGL was discovered experimentally and is not supported by theory, which in turn provides less information about the confidence of the model. Nonetheless, for a large proportion of applications the raw performance is the main goal to maximize. The search process did not cover the entire search space and in the future better performing functions may be found, thereby the NGL may point to a promising direction when searching for general suitable functions for classification tasks.

\section{Code}
Code can be found on the project page \url{https://github.com/ZKI-PH-ImageAnalysis/Next-Generation-Loss/tree/main}.

\section{Conclusions}
This study proposes to use Genetic Programming to search a generally applicable loss function for image classification tasks. During that process, a new function was found and shown to outperform other losses, commonly used for classification and segmentation tasks, on average on a variety of datasets demonstrating its general applicability. Moreover, it was shown that proposed loss function can be applied to train a variety of model architectures. Further analysis suggested that improvements provided by the new loss result from implicit regularization that reduces overfitting to the data.

{\small
\bibliographystyle{unsrtnat}
\bibliography{main}
}

\newpage
\appendix
\section{Method}
\begin{figure}[h!]
\renewcommand{\thefigure}{A}
\centering
\includegraphics[width=\textwidth]{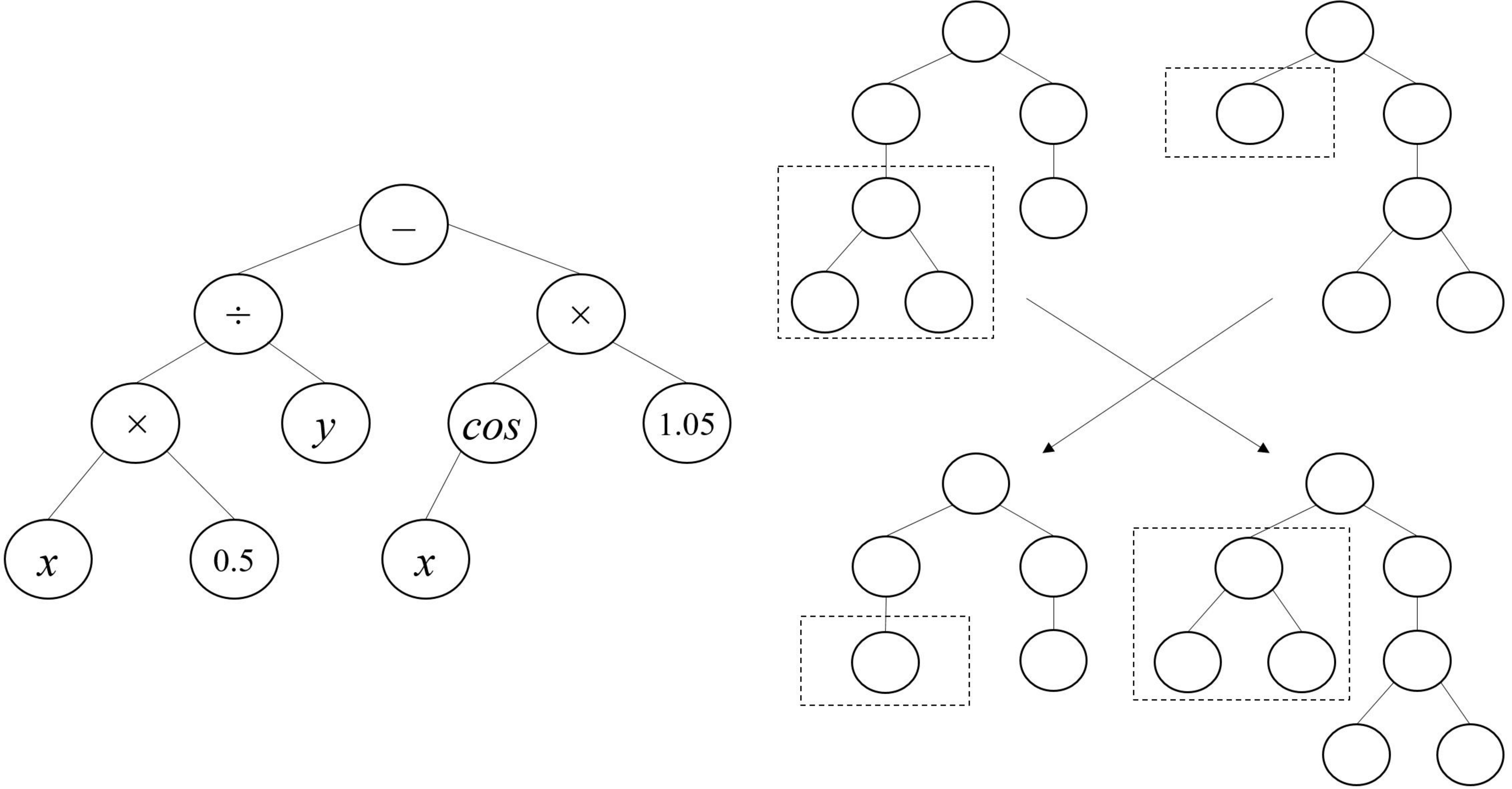}
\caption{Left: an example of the solution representation for tree-based GP: $\frac{0.5 \times x}{y} -1.05\times\cos(x)$. Right: an example of the subtrees exchange during crossover.} \label{fig_GP}
\end{figure}

The GP search process starts with random initialization of a set of potential solutions, which is also called a population of individuals, in the functional search space. For this study a set of loss functions was randomly generated, and each loss function was represented as a tree, where the terminals were randomly chosen from the set $\{y_{pred}, y_{real}, \mathbb{R}\}$, while the operators were chosen from the set $\{+, -, \times, \div, \times(-1), \sqrt, \log, \exp, \sin, \cos\}$. To utilize these functions and terminals for loss function generation, the following modifications were applied:
\begin{itemize}
\item $x \div y$ was changed to $x \div (y+\varepsilon)$;
\item $\sqrt{x}$ was changed to $\sqrt{|x|+\varepsilon}$;
\item $\log(x)$ was changed to $\log(|x|+\varepsilon)$, with $\varepsilon=10^{-8}$;
\item it was implemented as a rule that both $y_{pred}$ and $y_{real}$ should be present in the formula, and $y_{pred}\in[0,1]$ (Softmax is always the last layer of the model). 
\end{itemize}

\section{Experiments}
\subsection{Loss function search}

In this study, five experiments were conducted to search the loss function by GP, which only differed in how the fitness function $F$ was defined. Specifically, the following definitions of $F$ were used for the respective experiments:
\begin{itemize}
\item each individual loss function was used to train the InceptionV3 model from scratch one time on CIFAR-10 dataset, the validation error was used as the fitness value;
\item each individual loss function was used to train the InceptionV3 model from scratch 3 times on Fashion-MNIST dataset, the averaged validation error was used as the fitness value;
\item each individual loss function was used to fine-tune the pre-trained InceptionV3 model one time on CIFAR-10 dataset, the validation error was used as the fitness value;
\item each individual loss function was used to fine-tune the pre-trained InceptionV3 model 3 times on Fashion-MNIST dataset, the averaged validation error was used as the fitness value;
\item each individual loss function was used to train the InceptionV3 model from scratch on CIFAR-10, Fashion-MNIST and CIFAR-100 datasets (once each). The validation error for each dataset was compared to the respective validation error obtained by the same model trained using CE loss, and individual fitness was represented by the pair of numbers indicating the number of wins and the percentage of improvement compared to CE.
\end{itemize}

For the first four experiments, the classification error was used to evaluate individual's fitness, and, therefore, the individual with the smallest classification error was considered as the best one. However, for the last experiment, individuals were compared in the following way:
\begin{itemize}
\item if individual $f_i$ outperformed CE at least once and individual $f_j$ showed worse results on all 3 datasets, then $f_i$ was considered as the better loss function;
\item if both individuals $f_i$ and $f_j$ outperformed CE the same number of times, then the mean percentage of improvement for all winning datasets was calculated and the loss with higher mean value was considered as the better one;
\item if both individuals $f_i$ and $f_j$ showed worse results on all datasets compared to the CE loss, then the one with lower mean second value was considered as the better loss.
\end{itemize}

The fitness function $F$ evaluation procedure for all 5 experiments is described in Algorithm~\ref{alg_2}. It should be noted that in Algorithm~\ref{alg_2} the number of datasets was set to 1 for the first 4 experiments and 3 for the last one. Besides, the number of runs was equal to 1 for the first, third and last experiments, while it was set to 3 for the remaining experiments. 
\begin{algorithm}
\renewcommand{\thealgorithm}{B.1}
\caption{Fitness function $F$ evaluation}\label{alg_2}
\begin{algorithmic}
\Require Test error values of the InceptionV3 model trained by using CE on all given datasets
\For{each individual from population}
\For{each dataset used for a given experiment}
\For{each run}
\State Randomly generate train, val and test sets
\State Create an InceptionV3 model
\State Train the model for a given number of epochs
\State Evaluate model's accuracy on the test set
\If{only one dataset is used for an experiment}
\State Save model's test error
\Else
\State Determine whether individual outperformed CE or not
\State Save the percentage of improvement compared to CE
\EndIf
\EndFor
\If{only one dataset is used for an experiment}
\State Save individual's test error averaged over all runs
\Else
\State Calculate individual's number of wins compared to CE
\State Calculate the average percentage of improvement compared to CE
\EndIf
\EndFor
\If{only one dataset is used for an experiment}
\State Save individual's averaged test error
\Else
\State Save the number of wins and average percentage of improvement as a pair of values
\EndIf
\EndFor
\end{algorithmic}
\end{algorithm}

The best performing functions for each of the five GP experiments are listed below:
\begin{equation}
\label{eqn:eqA1}
f_1=\frac{1}{N}\sum_{i=1}^{N}{\frac{e^{(\sin^2(y^{(i)}_{pred}))}}{2\times(|(\beta-y^{(i)}_{real})\times y^{(i)}_{pred}| + \varepsilon)}},\tag{B.1.1}
\end{equation}
\begin{equation}
\label{eqn:eqA2}
f_2=\frac{1}{N}\sum_{i=1}^{N}{\left[\frac{\cos(-y^{(i)}_{real})}{\sqrt{\left|y^{(i)}_{pred}\right|+\varepsilon}+\varepsilon}\times(y^{(i)}_{pred}-\gamma)+(y^{(i)}_{pred})^4+\sqrt{\left|y^{(i)}_{real}+y^{(i)}_{pred}\right|+\varepsilon}\right]},\tag{B.1.2}
\end{equation}
\begin{equation}
\label{eqn:eqA3}
f_3=\frac{1}{N}\sum_{i=1}^{N}{\sqrt{\left|\frac{y^{(i)}_{real}}{\sin(y^{(i)}_{real})+\varepsilon}+\frac{y^{(i)}_{pred}-e^{\delta}}{\zeta}\right|+\varepsilon}},\tag{B.1.3}
\end{equation}
\begin{equation}
\label{eqn:eqA4}
f_4=\frac{1}{N}\sum_{i=1}^{N}{\sin(\sin(y^{(i)}_{pred}-\eta)+(y^{(i)}_{real})^3)},\tag{B.1.4}
\end{equation}
\begin{equation}
\label{eqn:eqA5}
f_5=NGL=\frac{1}{N}\sum_{i=1}^{N}{\left[e^{(\alpha-y^{(i)}_{pred}-y^{(i)}_{pred}\times y^{(i)}_{real})}-\cos(\cos(\sin(y^{(i)}_{pred})))\right]},\tag{B.1.5}
\end{equation}
where $\alpha=2.4092$, $\beta=1.5494$, $\gamma=3.8235$, $\delta=3.1868$, $\zeta=2.4428$, $\eta=2.6085$, $\varepsilon=10^{-8}$ and $N$ is the number of classes. Here in all formulas $y_{pred}=(y^{(1)}_{pred}, ..., y^{(N)}_{pred})$ is obtained after applying softmax function to the generated prediction, thus, $y^{(i)}_{pred}\in[0,1], i\in[1,N]$. Later in the study $f_5$ was denoted as $NGL$. 

\subsection{Small datasets}
Parameter settings and data preprocessing varied for all datasets but were identical for all tested loss functions, specifics are shown in Table~\ref{tab_A1}. Classification accuracy on the test set after the last epoch was used as the evaluation metric. 

\begin{table}
\renewcommand{\thetable}{B.2.1}
\caption{Parameter settings for used small datasets.}\label{tab_A1}
\resizebox{\textwidth}{!}{\begin{tabular}{cccccccc}
\toprule
& CIFAR-10 & Fashion-MNIST & CIFAR-100 & Malaria & PCam & Caltech 101 & Colorectal Histology \\
\midrule
Epochs & $200$ & $100$ & $200$ & $400$ & $50$ & $400$ & $600$\\
Optimizer & \multicolumn{4}{c}{Adam: $\beta_1 = 0.9$, $\beta_2 = 0.999$, $\epsilon = 10^{-7}$} & \multicolumn{3}{c}{SGD: momemtum $0.9$}\\
Learning rate $lr$ & \multicolumn{4}{c}{$0.001$} & \multicolumn{3}{c}{$0.01$}\\
Scheduler & \multicolumn{4}{c}{Reduce on Plateau: factor $0.2$, patience $5$, $lr_{min}=10^{-4}$} & \multicolumn{3}{c}{-}\\
Batch size & \multicolumn{3}{c}{$128$} & $32$ & $128$ & \multicolumn{2}{c}{$32$}\\
Image size & $32\times32$ & \multicolumn{2}{c}{$224\times224$} & $75\times75$ & $96\times96$ & $224\times224$ & $150\times150$\\
Augmentation & \multicolumn{7}{c}{Zoom $0.2$, width and height shift ($0.1$ for both), horizontral flip}\\
\bottomrule
\end{tabular}}
\end{table}

The evaluation of the five found functions was performed by training ResNet50 and InceptionV3 on seven datasets, which differed by the number of images, classes, by the type of images (grayscale and RGB), and their sizes. Classification accuracy on the test set after the last epoch was used as the evaluation metric. Both InceptionV3 and ResNet50 models were trained five times for each dataset. Obtained results were averaged for both models and compared to the averaged CE loss for the same models. In the end, the percentage of accuracy improvement or decrease compared to CE was determined. A final value from the range $[-0.5, 0.5]$ was considered as same performing, a value $>+0.5 \%$ better performing and $<-0.5 \%$ performing worse compared to CE. Here results obtained for all five evaluated loss functions ($f_1,...,f_5$) used to fine-tune pre-trained ResNet50 and InceptionV3 models are given in Table~\ref{tab_A2}.

\begin{table}
\renewcommand{\thetable}{B.2.2}
\caption{Average accuracy (\%) for pre-trained ResNet50 and InceptionV3 each fine-tuned 5 times for each dataset for cross entropy (CE) and percentual change of the tested loss functions compared to CE, best result per dataset is shown in bold.}\label{tab_A2}
\resizebox{\textwidth}{!}{\begin{tabular}{ccccccccc}
\toprule
Loss & Malaria & PCam & Colorectal Histology & CIFAR-10 & Fashion-MNIST & CIFAR-100 & Caltech 101 & Mean \\
\midrule
CE & 96.7 & 72.4 & 90.5 & 94.9 & 94.6 & 74.8 & 84.5 & $\pm 0$ \\
SCE & +0.11 & -0.83 & -2.13 & -2.12 & -1.82 & -1.25 & -0.02 & -1.13\\
Focal & +0.15 & +3.86 & +2.44 & +0.35 & +0.08 & -3.18 & +0.52 & +0.58\\
CE + $L2$ & \textbf{+0.34} & +0.73 & -2.77 & +0.50 & -0.08 & -0.23 & -1.19 & -0.39\\
$f_1$ &+0.20 &+7.90 &\textbf{+5.33} &-1.33 &-0.17 &+2.89 &-0.33 &+2.06\\
$f_2$ &-0.09 &\textbf{+8.62} &+4.54 &+0.19 &\textbf{+0.09} &+4.79 &+0.10 &+2.61\\
$f_3$ &-2.13 &+0.51 &-50.96 &\textbf{+1.74} &-0.04 &+4.98 &\textbf{+5.59} &-5.76\\
$f_4$ &-0.04 &+8.10 &+3.03 &+0.02 &-0.13 &\textbf{+6.78} &+5.47 &\textbf{+3.33}\\
$f_5 (NGL)$ &-0.26 &+7.07 &+5.01 &-0.37 &-0.30 &+5.29 &+3.45 &+2.84\\
\bottomrule
\end{tabular}}
\end{table}

\section{Discussion}
All functions found by GP and listed in Section 1.1 were analysed the same way as was done in~\cite{Gonzalez2019ImprovedTS}, thus, they were analysed in the context of binary classification to show why and how they work. For example, let us consider loss function $NGL$ for classification problem with $N=2$ classes:

\begin{equation}
\label{eqn:eqA6}
\begin{split}
NGL = \frac{1}{2}\left[e^{(\alpha-y_{pred}^{(0)} \cdot (1 + y_{real}^{(0)}))} + e^{(\alpha-y_{pred}^{(1)} \cdot (1 + y_{real}^{(1)}))}\right] - \\
- \frac{1}{2}\left[\cos{(\cos{(\sin{(y_{pred}^{(0)})})})} + \cos{(\cos{(\sin{(y_{pred}^{(1)})})})}\right] ,
\end{split}
\tag{B.2.1}
\end{equation}
where $\alpha=2.4092$.

Since vectors $y_{pred}$ and $y_{real}$ sum to $1$, by consequence of being passed through a softmax function, for binary classification $y_{pred} = (y_{pred}^{(0)}, 1-y_{pred}^{(0)})$ and $y_{real} = (y_{real}^{(0)}, 1-y_{real}^{(0)})$. This constraint simplifies the binary $NGL$ loss to the following function of two variables $(y_{pred}^{(0)}, y_{real}^{(0)})$:

\begin{equation}
\label{eqn:eqA7}
\begin{split}
NGL = \frac{1}{2}\left[e^{(\alpha-y_{pred}^{(0)} \cdot (1 + y_{real}^{(0)}))} + e^{(\alpha-(1-y_{pred}^{(0)}) \cdot (2 - y_{real}^{(0)}))}\right] - \\
- \frac{1}{2}\left[\cos{(\cos{(\sin{(y_{pred}^{(0)})})})} + \cos{(\cos{(\sin{(1 - y_{pred}^{(0)})})})}\right] ,
\end{split}
\tag{B.2.2}
\end{equation}
where $\alpha=2.4092$. This same methodology can be applied to the cross-entropy loss and other functions found by GP and listed in Section 2.1. 

\begin{figure}[t]
\renewcommand{\thefigure}{B.2}
\centering
\includegraphics[width=\textwidth]{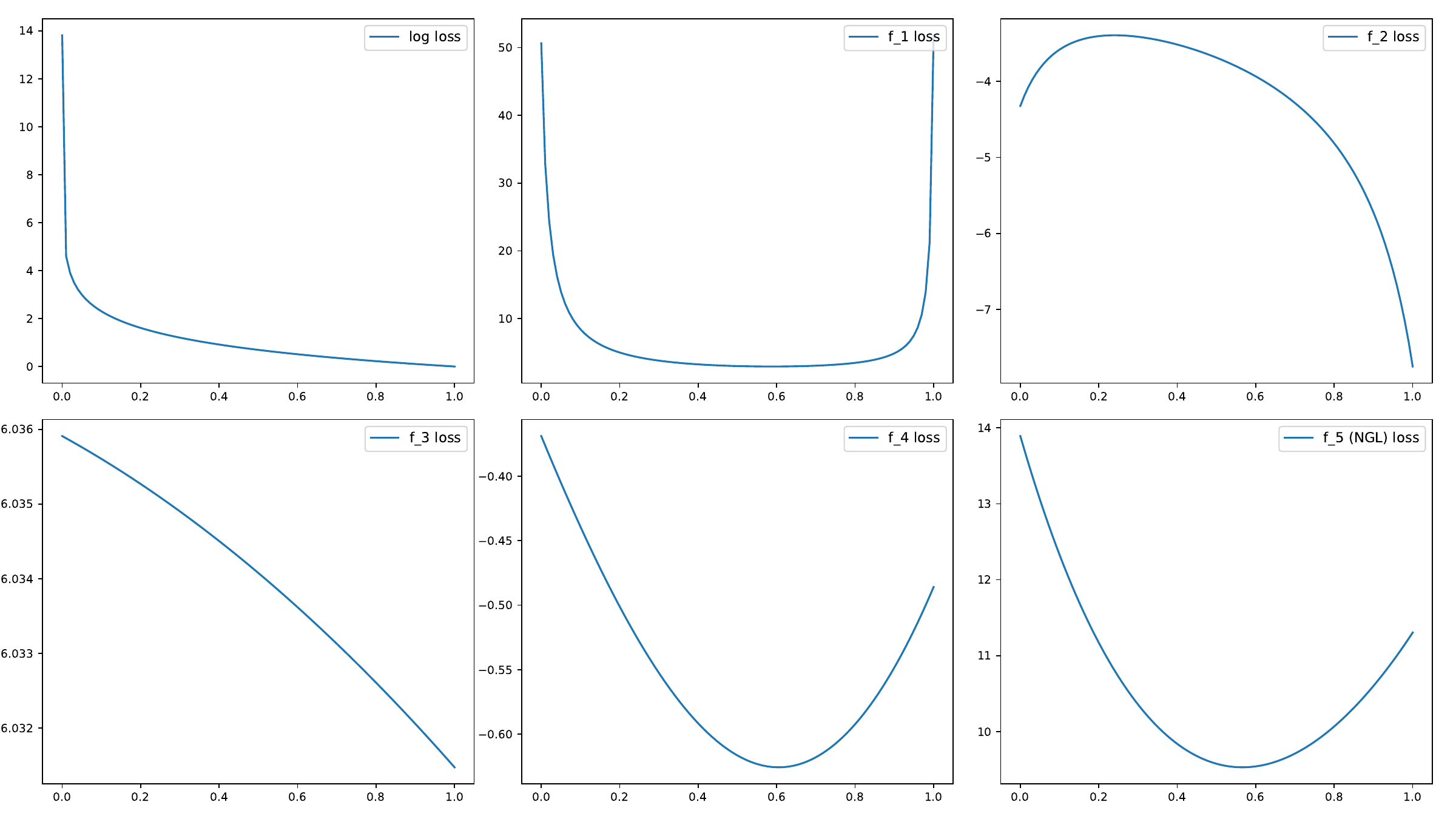}
\caption{Functions found by GP and listed in Section 2.1} \label{fig_loss_all_graphs}
\end{figure}

Let us assume that true labels $y_{real}$ are either $0$ or $1$. The case where $y_{real}=1$ is plotted in Fig.~\ref{fig_loss_all_graphs} for the CE loss and loss functions found by GP. The cross-entropy and $f_3$ losses are shown to be monotonically decreasing, $f_2$ loss has a small region where it increases and after that, similar to CE and $f_3$, it starts to decrease, while the remaining three functions, $f_1$, $f_4$ and $NGL$, show an increase in the loss value as the predicted label, $y_{pred}$, approaches the true label $y_{true}$. This increase of the loss value allows the loss functions $f_1$, $f_4$ and $NGL$ to prevent the model from becoming too confident in its output predictions. The latter may provide an important advantage as it lowers the probability of network's overfitting. Thus, these loss functions provide an implicit form of regularization, enabling better generalization.

Results, achieved on small datasets, show that the only found function, which monotonically decreases, $f_3$, demonstrates worst results as for models trained from scratch and for pretrained models. Loss $f_2$ shows almost the same results as CE on most of small datasets, but is significantly worse on datasets such as Colorectal Histology and Caltech101. Despite the fact that generally results for $f_1$ are good, it should be noted that function $f_1$ (similarly to $f_2$) depends on the $\varepsilon$ value, which has significant impact on the value of $f_1$ itself and makes the latter differentiable. Therefore, only functions $f_4$ and $NGL$ have the mentioned form of regularization and do not depend on additional parameters such as introduced $\varepsilon$ value. It can be seen from Fig.~\ref{fig_loss_all_graphs}, that the minimum for the $f_4$ loss, where $y_{real}=1$, lies near $0.61$, and the minimum for the $NGL$ loss is near $0.57$ when $y_{real}=1$. All losses were used to train ResNet101 model on ImageNet-1k dataset, however, experiments showed that only $NGL$ showed reasonable performance. 

Thus, loss $NGL$ was chosen for the experiments on large datasets, such as ImageNet-1k and COCO-Stuff164k, as it doesn't depend on any additional parameters such as $\varepsilon$ to make it diffrentiable, has implicit regularization and showed the best results on small datasets (pre-trained and trained from scratch). Experiments on ImageNet-1k and COCO-Stuff164k showed that found loss function $NGL$ is "slower" compared to CE, for example, as the latter is capable to reach generally good accuracy very quickly. Nevertheless, given the same number of epochs, $NGL$ loss manages to catch up and surpass CE later during the training. It can be caused by the fact that at some point during the training cross-entropy loss stops or barely improves deep learning model's accuracy, while $NGL$ function is more "cautious" during the training due to its implicit regularization, which helps model to prevent overfitting.

\end{document}

\typeout{get arXiv to do 4 passes: Label(s) may have changed. Rerun}